\documentclass[10pt,conference,a4paper]{IEEEtran}

\IEEEoverridecommandlockouts
\usepackage{cite}
\usepackage{amsmath,amssymb,amsfonts}
\usepackage{algorithmic}
\usepackage{graphicx}
\usepackage{textcomp}
\usepackage{xcolor}
\usepackage{comment}
\usepackage{multirow}
\usepackage{subfig}
\usepackage{algorithm,algorithmic}

\newcommand{\argmin}{\mathop{\rm arg~min}\limits}

\usepackage{color}

\def\BibTeX{{\rm B\kern-.05em{\sc i\kern-.025em b}\kern-.08em
    T\kern-.1667em\lower.7ex\hbox{E}\kern-.125emX}}
\begin{document}

\title{Filter Pruning using Hierarchical Group Sparse Regularization for Deep Convolutional Neural Networks
\thanks{This work was partly supported by JSPS KAKENHI Grant Number 16K00239.}
}

\author{\IEEEauthorblockN{Kakeru Mitsuno}
\IEEEauthorblockA{\textit{School of Engineering} \\
\textit{Hiroshima University}\\
Higashi Hiroshima, Japan \\
mitsunokakeru@gmail.com
}

\and
\IEEEauthorblockN{Takio Kurita}
\IEEEauthorblockA{\textit{Department of Information Engineering} \\
\textit{Hiroshima University}\\
Higashi Hiroshima, Japan \\
tkurita@hiroshima-u.ac.jp}
}

\maketitle

\begin{abstract}
Since the convolutional neural networks are often trained with redundant parameters, it is possible to reduce redundant kernels or filters to obtain a compact network without dropping the classification accuracy.
In this paper, we propose a filter pruning method using the hierarchical group sparse regularization. 
It is shown in our previous work that the hierarchical group sparse regularization is effective in obtaining sparse networks in which filters connected to unnecessary channels are automatically close to zero.
After training the convolutional neural network with the hierarchical group sparse regularization, the unnecessary filters are selected based on the increase of the classification loss of the randomly selected training samples to obtain a compact network.
It is shown that the proposed method can reduce more than 50\% parameters of ResNet for CIFAR-10 with only 0.3\% decrease in the accuracy of test samples.
Also, 34\% parameters of ResNet are reduced for TinyImageNet-200  with higher accuracy than the baseline network.

\end{abstract}


\section{Introduction}

\begin{figure}[tbp]
\centering
\subfloat[Sparse kernels of layer $l+1$ which are connected to the unnecessary output channel of layer $l$]{\includegraphics[scale=0.25]{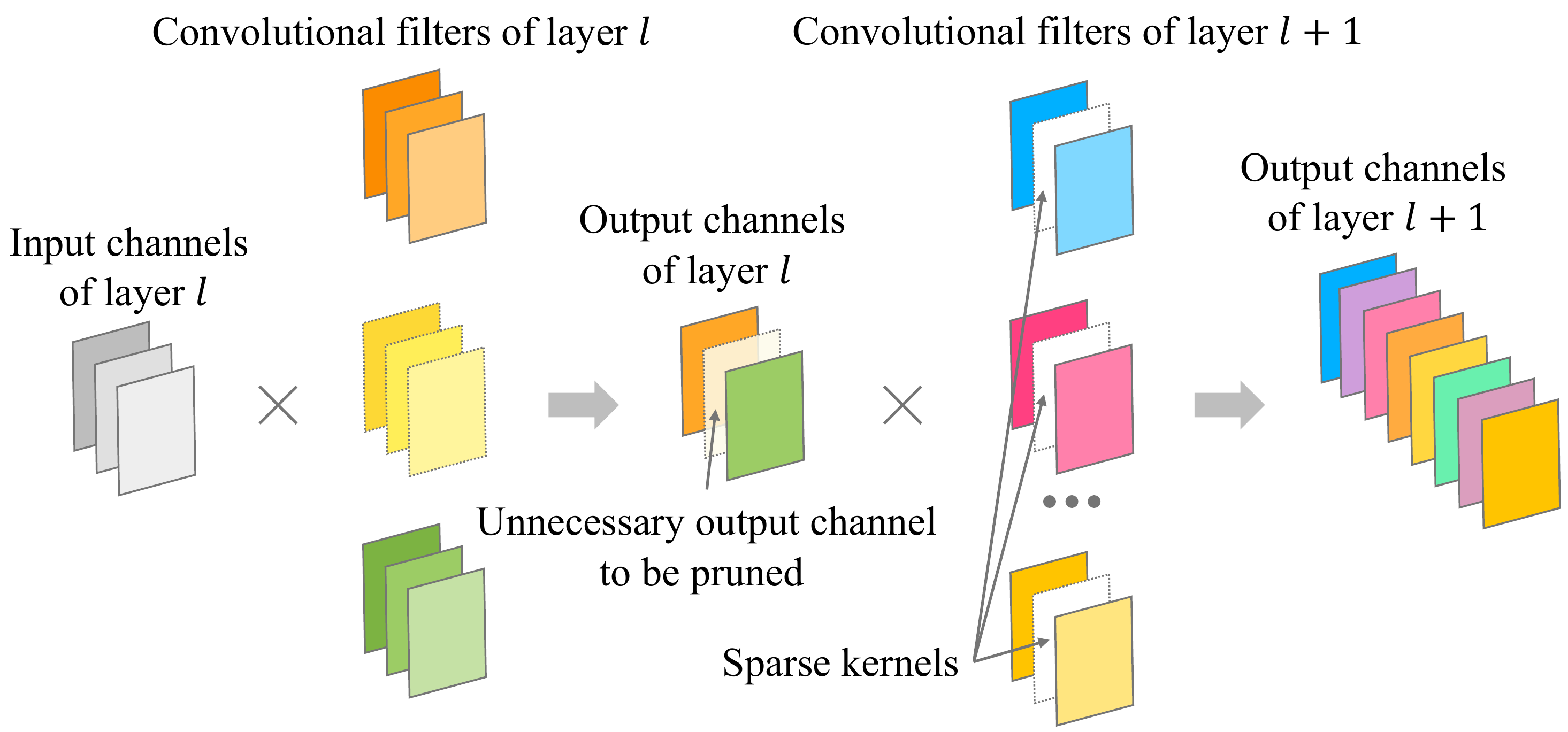}
\label{filterpruning1}}
\hfil
\subfloat[Pruning the filter of layer $l$ and sparse kernels of layer $l+1$, which are connected to the unnecessary output channel of layer $l$ keeping the same output channel of layer $l+1$]{\includegraphics[scale=0.25]{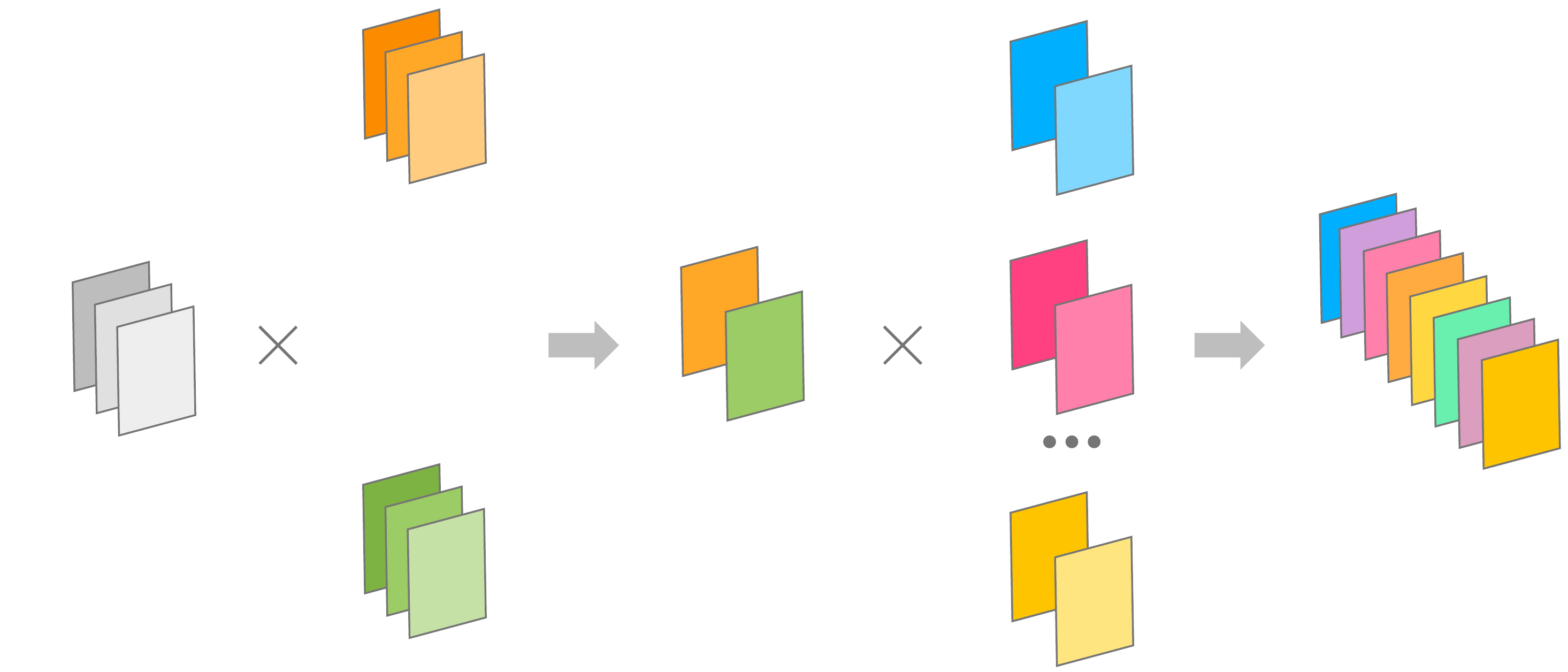}
\label{filterpruning2}}
\caption[]{An illustration of filter pruning via Hierarchical sparse group regularization based on the feature-wise grouping. 
In convolutional layer, each filter makes one output channel (activation), these colors are the same.
For example, the filter of orange of layer $l$ makes orange output channel of layer $l$.
\subref{filterpruning1} The Hierarchical sparse group regularization based on the feature-wise grouping make the weights of the unnecessary kernels to be almost zero. 
Since the output of convolution from the input channel connected to the unnecessary kernels will be zero in the layer $l+1$, the output channels are not influenced by the pruning the unnecessary kernel.
\subref{filterpruning2}
If the increase of the classification loss of the network after pruning the filters connected to the unnecessary output channel of layer $l$ is very small, we can prune the filters of layer $l$ and kernels of layer $l+1$ connected to the unnecessary output channel of layer $l$.
Then the output channels of layer $l+1$ are almost the same as the output channels before pruning.
}
\label{FPHGSR}
\end{figure}

Deep convolutional neural networks (CNNs) have been successful with excellent performance in computer vision tasks such as image classification \cite{krizhevsky2012imagenet, szegedy2015going}. But along with it, the network becomes deeper and wider, requiring excessive parameters\cite{he2016deep,huang2017densely}, which increases the computational cost. To solve this problem, many approaches have been proposed such as knowledge distillation \cite{hinton2015distilling, romero2014fitnets} and network pruning \cite{han2015learning, liu2017learning, he2019filter}, that can be reduced redundant parameters of large networks while preserving accuracy.

Recent work on network pruning, there are 2 types of methods by using sparse regularization to obtain a sparse network whose unnecessary connections are pruned. 
One of the methods is unstructured pruning \cite{han2015learning,louizos2017learning}, which prune individual weights by using L0 regularization, L1 regularization\cite{tibshirani1996regression} or L2 regularization. 
These unstructured sparse regularizations enforce unnecessary individual weights of a large network set to be 0.
However, such unstructured pruning cannot prune whole kernels or 
filters (that is a subset of kernels connected to a channel) 
in CNNs, which does not speed up computations or reduce memory weight without special libraries and hardware. 

In contrast, structured pruning enables to reduce the number of kernels or filters from CNNs \cite{wen2016learning, alvarez2016learning, zhou2016less} by using group sparse regularization such as group lasso\cite{yuan2006model,schmidt2010graphical}, whose method is widely used to pruning unnecessary kernels or filters keeping high performance.

The weights of the convolutional layer are structured as convolution kernels. 
When the filters connected to a channel are considered as a group, group sparse regularization treats a subset of kernels as individual weight in the same group, mutual interaction between kernels in the group is not taken into account.
In our previous work, we proposed the concept of the hierarchical group sparse regularization \cite{mitsuno2020hierarchical} to introduce such interactions in the structured sparse regularization criterion.
The hierarchical group sparse regularization can treat the filter for the input channel or the output channel as a group and convolutional kernels as a group in the same group to prune the unnecessary subsets of weights. As a result, we can prune the weights more adequately depending on the structure of the network and the number of channels keeping high performance.

In this paper, we propose a filter pruning method with the hierarchical group sparse regularization based on the feature-wise grouping 
as illustrated in Fig. \ref{FPHGSR}, 
whose grouping consider as a group the kernels connected to a input channel. By this grouping, we can prune unnecessary output channels in layer $l-1$ (the input channels in layer $l$).
After training with the hierarchical group sparse regularization, we calculate the influence of each channel on the classification loss of the randomly selected training samples, and prune the channels based on the increase of the classification low. 
After we obtain a compact network by the filter pruning, the parameters of the compact network are retrained from scratch.


To confirm the effectiveness of the proposed method, we have performed experiments with different network architectures (VGG and ResNet) on different data sets (CIFAR-10, CIFAR-100, and TinyImageNet-200).
The results show that we can obtain the compact networks with about 50\% less parameters without decrease of the classification accuracy.


The contributions of this paper are summarized as follows:
\begin{itemize}
  \item We propose a filter pruning method with the hierarchical group sparse regularization based on the feature-wise grouping, the regularization can prune filters more adequately depending on the structure of the network and the number of channels than non-hierarchical sparse regularization.
  \item The feature-wise grouping can prune the filters connected to unnecessary input channels by removing the channels with low influence on the classification loss.
  \item The effectiveness of the proposed pruning method is confirmed through experiments with different network architectures (VGG and ResNet) on different data sets (CIFAR-10, CIFAR-100, and TinyImageNet-200). 
\end{itemize}

\section{Related Work}

\subsection{Network Pruning}

Network pruning can efficiently prune redundant weights or filters to compress deep CNN while maintaining accuracy. Various method have been proposed.
There is two methods of network pruning in which unstructured pruning and structured pruning. 

\subsubsection{Unstructured Pruning}
The way of unstructured pruning reduces the individual weight of neural networks \cite{lecun1990optimal,hassibi1993second,han2015learning,han2015deep,srinivas2015data,louizos2017learning}. 
Optimal brain damage \cite{lecun1990optimal} and optimal brain surgeon \cite{hassibi1993second} prune unimportant weight from a network to compute the influence of each weight on the training loss based on Hessian matrix. 
S.~Han et al. \cite{han2015learning} and C.~Louizos et al.\cite{louizos2017learning} utilized unstructured sparse regularizations such as L1 regularization to make a sparse networks, which reduce unnecessarily individual weights by enforcing to be 0. 
These unstructured pruning method makes network weights sparse, but it can only achieve speedup and compression with dedicated libraries and hardware.

\subsubsection{Structured Pruning}

Structured pruning \cite{li2016pruning,hu2016network,wen2016learning, alvarez2016learning, zhou2016less,he2017channel,molchanov2016pruning,peng2019collaborative,molchanov2019importance,liu2017learning,ye2018rethinking,huang2018data,luo2017thinet,yu2018nisp,suau2018principal,huang2018learning,chin2018layer,lin2018accelerating,he2018soft,wang2018skipnet,liu2018rethinking,peng2019collaborative,lin2019towards,he2019filter} has been exploited by many researchers.
Pruning methods using group lasso as structured sparse regularization were proposed by many researchers \cite{wen2016learning, alvarez2016learning, zhou2016less}. 
These methods enforce unnecessary subset of weights to be zero to obtain sparse networks at group levels such as kernels or filters.
Also, Y.~He et al. \cite{he2017channel} proposed channels selection based on lasso regression.
Pruning of unnecessary filters with the scaling parameter of batch normalization layers were also proposed by enforcing sparsity of the parameter \cite{liu2017learning,ye2018rethinking,huang2018data}.
In the papers 
\cite{li2016pruning,he2018soft}, filters with relatively low weight magnitudes were pruned based on the norm of the weights.
Y.~He et al. \cite{he2019filter} proposed to prune the most replaceable filters containing redundant information and the relatively less important filters using the norm-based criterion.
Some works \cite{molchanov2016pruning,lin2018accelerating,peng2019collaborative,molchanov2019importance} used Taylor expansions to evaluate the influence of pruned filters to the classification loss.

\subsection{Sparse Regularization}
In this section, we review structured sparse regularization criteria for filter pruning of deep neural networks.

We assume that the objective function of the optimization for determining the trainable weights is given by
\begin{equation}
\label{training objective}
J(W) = \mathbb{L}(f(x,W)|y) + \lambda \sum^{L}_{l=1}R(W^l)
\end{equation}
where $(x,y)$ denotes the pair of the input and target, $W$ is a set of all trainable weights of all the $L$ layers in the CNN,  $\mathbb{L}(\cdot)$ is the standard loss for the CNN, and $R(W^l)$ is the regularization term at layer $l$.
The parameter $\lambda$ is used to balances the loss and the regularization term.

Also, we assume the weight in the layer $l$ as $W^l \in \mathbb{R}^{C_l\times C_{l-1}\times K_l \times K_l}$, where $C_l$ and $C_{l-1}$ are the number of output channels and input channels, $K_l$ is the kernel size of the layer $l$ respectively. In the fully connected layers, $K_l = 1$

The structured sparse regularization such as group lasso regularization\cite{yuan2006model,schmidt2010graphical}, exclusive sparsity \cite{zhou2010exclusive,kong2014exclusive} and group $L_{1/2}$ regularization criterion \cite{li2018smooth,alemu2019group} can reduce unnecessary kernels or filters on CNN. 
The structured sparse regularization is defined as
\begin{equation}
\label{structured sparse regularization}
R(W^l) = \sum_{g \in G} r(W^l_g),
\end{equation}
where $g \in G$ is a group in the set of groups $G$, $W^l_g$ is subset of weights for the group $g$ such as a filter that connected to a channel, and $r(\cdot)$ is the function to calculate the sparseness of the structured sparse regularization. 
In the case of group lasso, the calculate of the regularization is written as $r(W^l_g)=\|W^l_g\|_2$.

\begin{figure}[tbp]
\centering
\subfloat[the neuron-wise]{\includegraphics[scale=0.2]{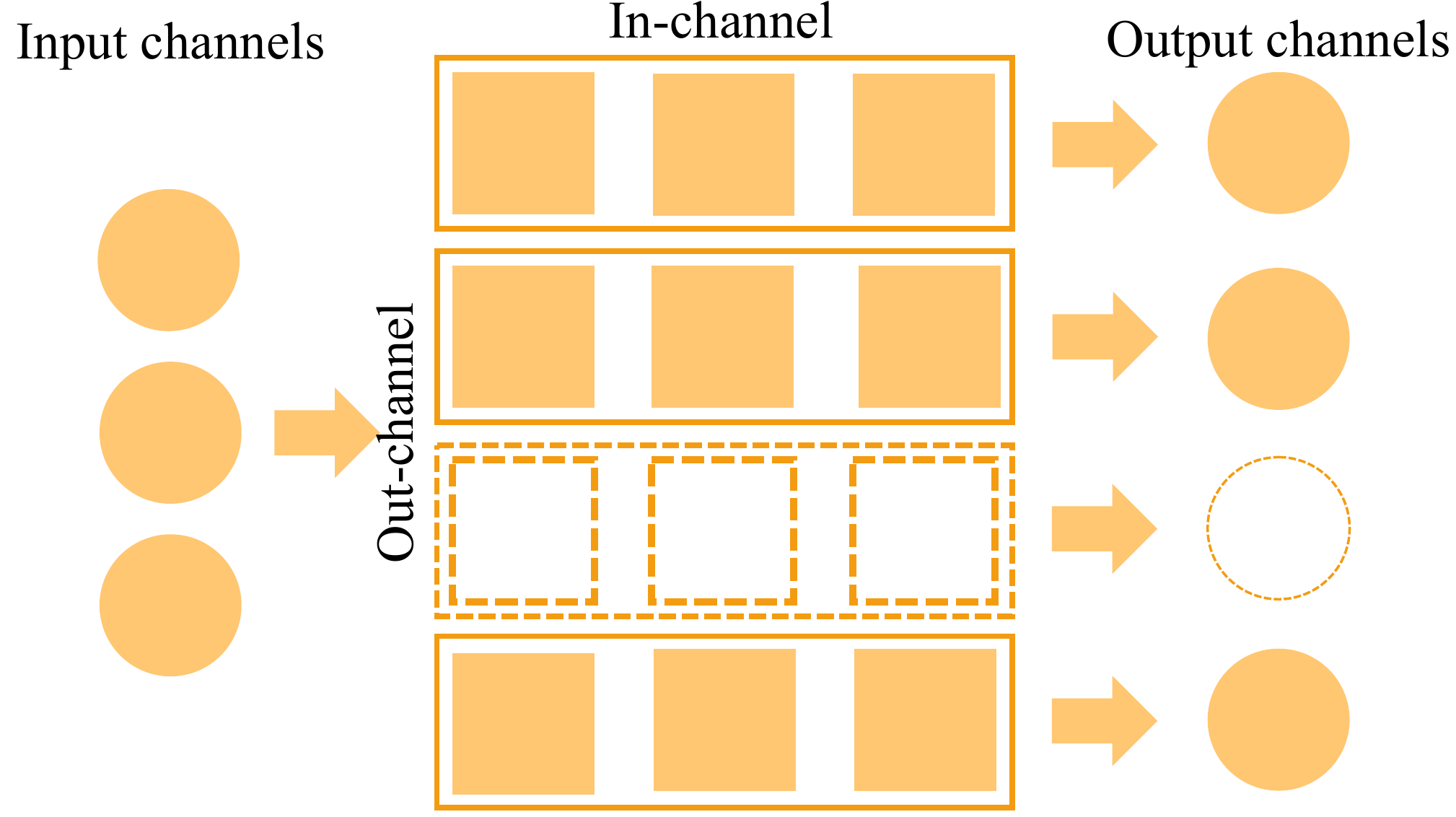}
\label{neuron-wise grouping}}
\hfil
\subfloat[the feature-wise]{\includegraphics[scale=0.2]{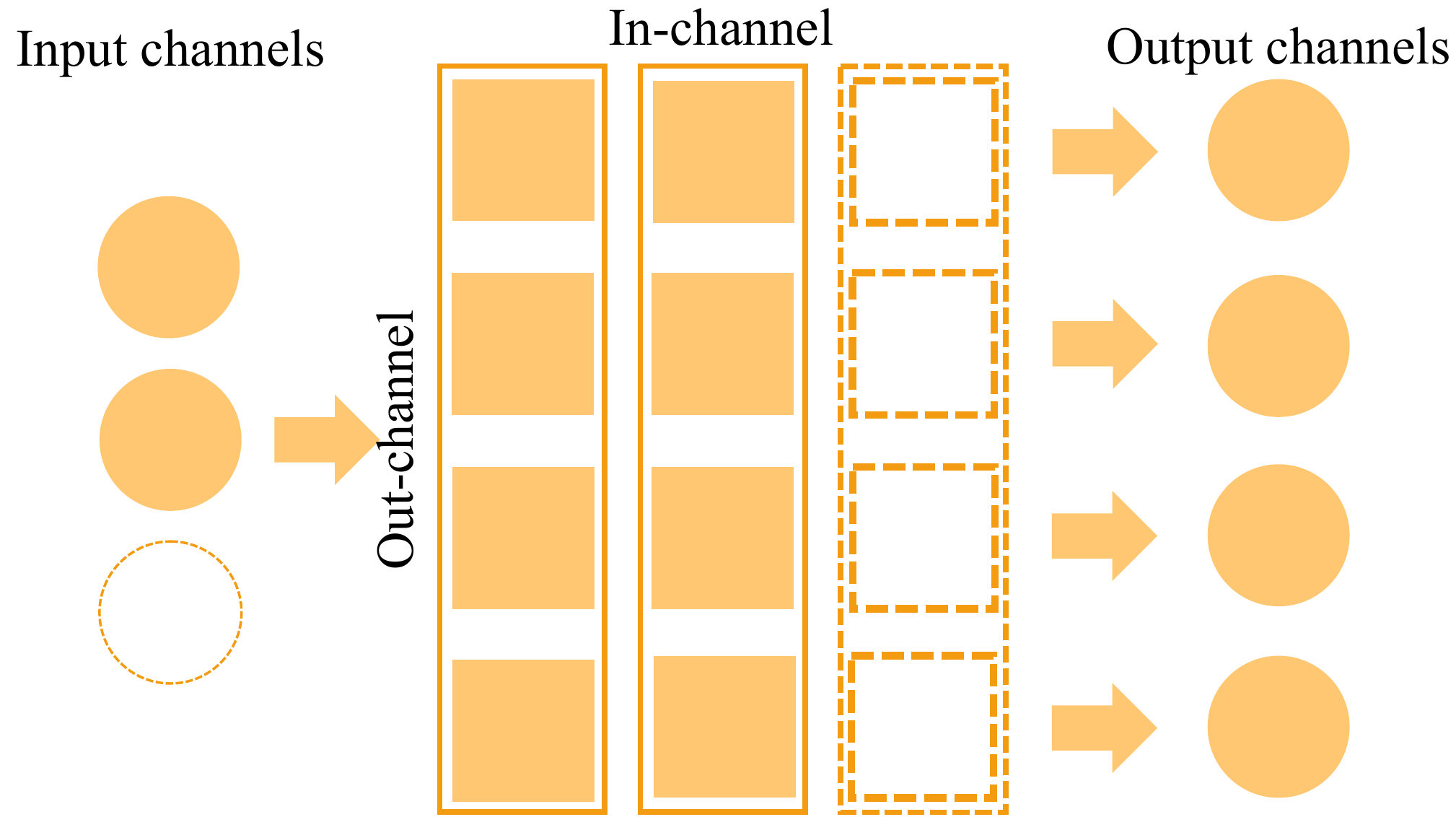}
\label{feature-wise grouping}}
\caption[]{The way of grouping for convolutional kernels.
\subref{neuron-wise grouping} The filters connected to a output channel are consider as a group. We call this grouping the neuron-wise grouping. By this grouping, we can prune unnecessary output channels.
\subref{feature-wise grouping} The filters connected to a input channel are considered as a group. We call this grouping the feature-wise grouping. By this grouping, we can prune unnecessary of output channels in layer $l-1$ (the input channels in layer $l$).}
\label{the way of grouping}
\end{figure}

To prune filters that connected unnecessary channels, there are two types of grouping, namely the neuron-wise grouping and the feature-wise grouping. The way of grouping for convolutional kernels are shown in Fig.~\ref{the way of grouping}.

In these groupings, structured sparse regularization treats subset of kernels as individual weight in the same group.
So, mutual interaction between kernels in the group is not taken into account.
In our previous work \cite{mitsuno2020hierarchical}, we proposed the concept of the hierarchical group sparse regularization to introduce such interactions in the structured sparse regularization criterion.

For example, the hierarchical group sparse regularization criterion is defined by taking the square of the sub-groups as
\begin{equation}
\label{hierarchical group Lasso regularization square}
r_{SQ}(W^l_g) = \left(\sum_{k \in K}r(W^l_{g,k})\right)^2,
\end{equation}
where $k \in K$ is a kernel in a filter (a set of kernels) $K$ and $w^l_{g,k}$ is a kernel in the group $g$ and $r(\cdot)$ is non-hierarchical group sparse regularization such as group lasso, exclusive sparsity and group $L_{1/2}$.

\section{Proposed Method}


We propose the feature-wise filter pruning algorithm for deep convolutional neural networks.
It consists of the following steps;
1) Train the large network as the initial network.
2) Train the network with the structured sparse regularization based on the feature-wise grouping to find unnecessary filters connected to input channels by enforcing the weights of unnecessary filters to be zero.
3) Prune the filters with smaller influence on the classification loss.
4) Train the obtained compact network from scratch.
The details of these steps are explained in the next sub-sections.

\subsection{Training with The Hierarchical Sparse Regularization Based on The Feature-Wise Grouping}

First, we train the network as the initial network without sparse regularization.

After obtaining the initial network, we train the network with the structured sparse regularization based on the feature-wise grouping such as group lasso, exclusive sparsity, group L$_{1/2}$ regularization, and the hierarchical group sparse regularization to get the sparse network.
In this paper, we propose to use the hierarchical group sparse regularization, which is proposed by Mitsuno et al. \cite{mitsuno2020hierarchical}, and show that the hierarchical group sparse regularization performs better than the non-hierarchical regularization.

The feature-wise group sparse regularization is defined as
\begin{equation}
\label{feature-wise grouping criteria}
R(W^l) = \sum^{c_{l-1}}_{j=1}r(W^l_{,j,,}) .
\end{equation}
The structured sparse regularization criterion based on the feature-wise grouping enforces the filters connected to unnecessary input channels of the convolutional layer to be zero.
As a result, we can remove unnecessary filters connected to output channels in layer $l-1$ by forcing the unnecessary filters connected to the input channels in layer $l$ to be zero.
Since the unnecessary input channels connected to the filters don't have influences to the outputs of the layer and final loss of the network, we can prune the filters connected to the output channels of the layer $l-1$ (the unnecessary input channels of the layer $l$).


\subsection{Filter Pruning}

After training with the structured sparse regularization based on the feature-wise grouping, we obtain sparse network in which all weights of the filters connected to the unnecessary input channels are close to zero.
Thus we can prune the filters connected to the unnecessary output channels of layer $l-1$ (the unnecessary input channels of the layer $l$) if the filters have less influence to the classification loss after pruning.
We can implement the filter selection method for pruning as shown in Algorithm. \ref{Pruning selection algorithm}.

\begin{algorithm}
 \caption{Filter Pruning (backward filter selection)}
 \label{Pruning selection algorithm}
 \begin{algorithmic}[1]
 \renewcommand{\algorithmicrequire}{\textbf{Input:}}
 \renewcommand{\algorithmicensure}{\textbf{Output:}}
 \REQUIRE filter of the sparse model $W\in \mathbb{R}^{L}$, the number of conv layer $L^c$, the number of all output channel of the conv layer $C^c$, filter mask $m^{l}_{c^{l}} \in M^l$, training sample $s \in x$, loss $\mathbb{L}(\cdot)$, pruning rate $P$
 \ENSURE  weight of pruned model $W^*\in \mathbb{R}^{L}$
 \STATE $M = 1$
  \FOR {$n$ in 1 \ldots $C^{c}P$}
     \FOR {$l$ in 1 \ldots $L^{c}$}
        \FOR {$c$ in 1 \ldots $C^{c}_{l}$}
            \STATE $M^* = M$
            \STATE $m^{*l}_{c} = 0$
            \STATE $e_{l,c} = \mathbb{L}(f(s,W\odot M^*)|y)$
        \ENDFOR
    \ENDFOR
    \STATE $\argmin_{l^*\in L^{c},c^* \in C^{c}_{l}}$ $e_{l^{*},c^{*}}$ s.t. $m^{l^{*}}_{c^{*}} == 1$
    \STATE $m^{l^{*}}_{c^{*}} = 0$
  \ENDFOR
  \STATE pruning $W^* = W[M == 1]$
 \RETURN $W^*$ 
 \end{algorithmic} 
 \end{algorithm}

Molchanov et al. \cite{molchanov2016pruning} proposed to use a criteria based on Taylor expansion for ranking and pruning one filter at a time.
Chin et al. \cite{chin2018layer} also introduced a global ranking approach.

Here we use the global ranking for filter pruning to automatically obtain the pruned network architecture. 
We introduce a binary mask $m^{l}_{c^{l}} \in M^l $ for each output channel of the layer $l$, $m^{l}_{i} \in \{0,1\}$.
By using these masks, we can prune $i^{th}$ filter of the layer $l$ by making $m^{l}_{i} = 0$.
We search the filter which has the minimum loss increase after the filter is pruned. The increase of the classification loss after pruning is represented as
\begin{eqnarray}
  \argmin_{l^*\in L^{c},i^* \in C^{c}_{l}} |\mathbb{L}(f(s,W)|y) - \mathbb{L}(f(s,W\odot M)|y)| 
\end{eqnarray}
where $W\odot M$ denotes the element-wise product.
The classification loss $\mathbb{L}$ are evaluated by using randomly selected training samples.
By updating the mask and repeating the filter selection algorithm until the number of the masked channels are $C^{c}P$ channels. After the filters for pruning are selected, we prune the filters from the network, where $m^{l}_{i} = 0$. 
We also prune the batch normalization layers connected to the pruned output channels at the same time.
Then we can obtain the compact network architecture with small parameters and less computational operations. 

The obtained compact network after pruning can achieve almost the same accuracy with the original large network by fine-tuninig.
But Liu et al. \cite{liu2018rethinking} examined and showed that the fine-tuning of the pruned model can only give comparable or worse performance than the training of the compact model with randomly initialized weights.
So we trained the obtained compact network from scratch. 

\section{Experiments}

To confirm the effectiveness of the proposed pruning method, we have performed experiments using different data sets (CIFAR-10, CIFAR-100, and TinyImageNet-200) and different network architectures (VGG nets \cite{simonyan2014very} and ResNet \cite{he2016deep}).
CIFAR-10 contains 60,000 color images of ten different animals and vehicles.
They are divided into 50,000 training images and 10,000 testing images. 
The size of each image is $32 \times 32$ pixels. 
CIFAR-100 also contains 60,000 color images of 100 different categories and 50,000 images are used for training and the remainings are used for test.
The size of each image is also $32 \times 32$ pixels. 
TinyImageNet-200 contains 110,000 color images of 200 different categories and 100,000 images are used for training and 10,000 images for test. 
The size of the image is $64 \times 64$ pixels. 

In the following experiments, 
the number of channels of the network at each layer is adjusted to prevent overfitting, depending on each dataset.

\subsection{Pruning of VGG Nets}

To confirm the effectiveness of the proposed pruning method, we have performed experiments using VGG14 (13-conv + 1-fc layers) with batch normalization layers for different data sets (CIFAR-10, CIFAR-100, and TinyImageNet-200).


%
%
All the initial networks are trained from scratch by using SGD optimizer with a momentum of $0.9$.
We used the weight decay with the strength of $5*10^{-4}$ to prevent overfitting.
The mini-batch size for CIFAR-10/100 was set to $128$ and the network was trained for $200$ epochs. 
For TinyImageNet-200, the mini-batch size was set to $256$ and the network was trained for $200$ epochs. 
The initial learning rate was set to $0.1$ and it was divided by $0.2$ after $[60,120,160]$ training epochs.


Then the hierarchical sparse regularization was applied to the weights except for the bias term in all convolutional layers.
All the networks were trained by using SGD optimizer with a momentum of $0.9$.
For CIFAR-10/100, the mini-batch size was set to $128$ and the network was trained for $100$ epochs.
For TinyImageNet-200, the mini-batch size was set to $256$ and the network was trained for $100$ epochs. 
The initial learning rate of $0.01$ which is divided by $0.1$ after $1/3$ and $2/3$ training epochs.
The hyper-parameter $\lambda$, which balances the cross-entropy loss and the hierarchical sparse regularization criterion, was experimentally determined in the range from $10^{-1} $ to $10^{-7}$.

We used hierarchical squared GL$_{1/2}$ regularization (HSQ-GL12)\cite{mitsuno2020hierarchical}, which has shown outstanding performance compare to various sparse regularization criteria in our previous work.
To evaluate the sparsity of the trained network, the ratio of the zero weights was calculated by assuming that the weights whose absolute value is less than $10^{-3}$ are zero.

After training the networks with sparse regularization, we selected one best trained network from the trained networks with the various hyper-parameter $\lambda$ in sparse regularization.


To obtain the compact network, the channels at the convolutional layers except for the output channel of the final convolutional layer were pruned based on the influence of the pruning of each channel to the classification loss.
To evaluate the influence of the pruning, 
we calculate the increase of the classification loss for the $128$ randomly selected samples from the training samples.

The pruning rate $P$, which is a ratio of the pruned channels in the whole channels of the networks, was changed from $0.1$ to $0.9$. 
When the number of channels of a layer becomes $0$, the experiments are stopped.
After pruning, we can obtain compact networks. 
The parameters of the obtained compact network were trained from scratch using the same parameter settings as the training of the initial networks.


\begin{figure}[tbp]
\centering
\subfloat[VGG14 on CIFAR-10]{
\includegraphics[scale=0.28]{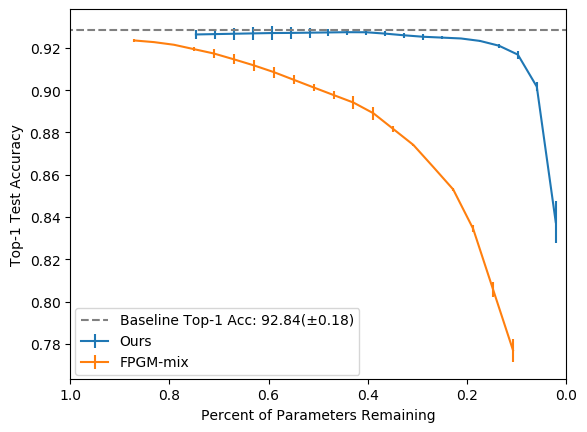}
\includegraphics[scale=0.28]{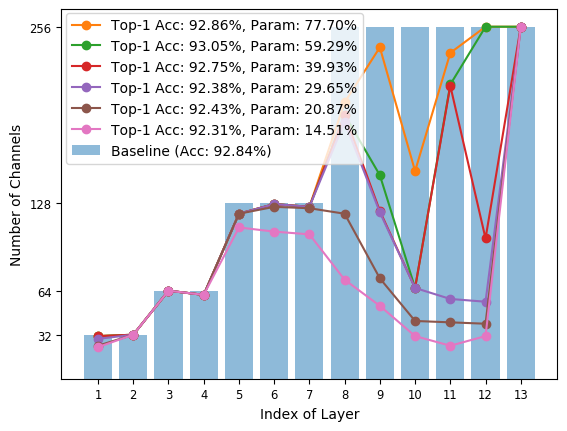}
\label{VGG14 on CIFAR-10}}
\hfil
\subfloat[VGG14 on CIFAR-100]
{
\includegraphics[scale=0.28]{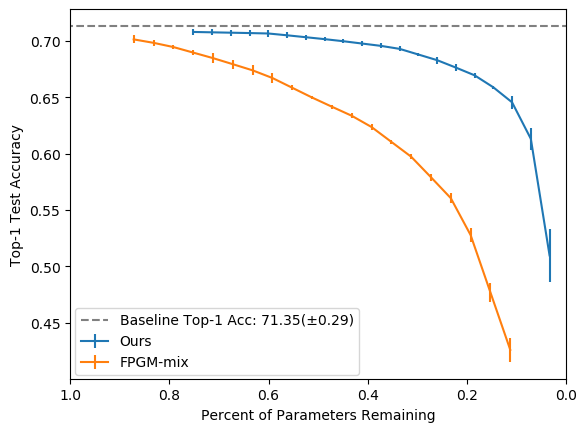}
\includegraphics[scale=0.28]{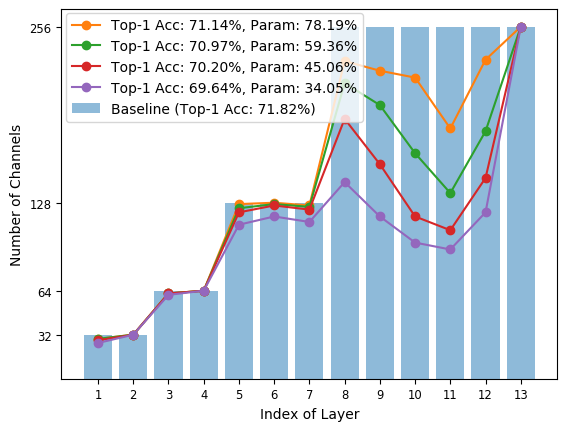}
\label{VGG14 on CIFAR-100}
}
\hfil
\subfloat[VGG14 on TinyImageNet-200]
{
\includegraphics[scale=0.28]{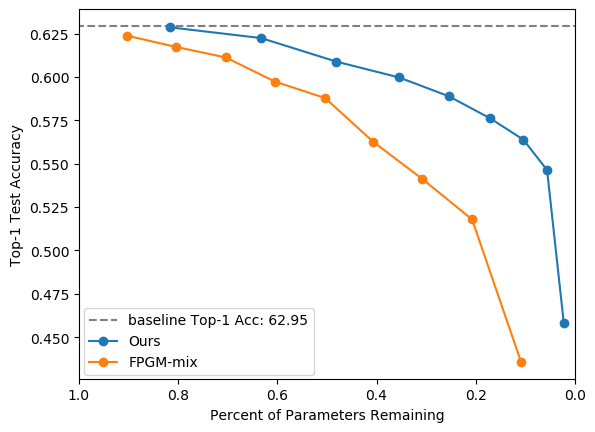}
\includegraphics[scale=0.28]{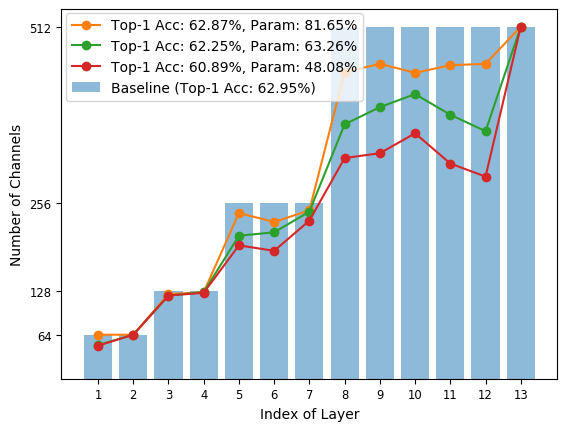}
\label{VGG14 on TinyImageNet-200}
}
\caption[]{The results with VGG14 on CIFAR-10/100 and TinyImageNet-200. Comparison of test accuracy of the pruned networks (left). For CIFAR-10/100, the average of three trials are shown. The numbers of the pruned channels in each layer is shown in the right figure. Each line shows the numbers of the pruned channels at each layer. Different colors denote the results with different pruning rates $P$.
}
\label{VGG nets}
\end{figure}

The results for VGG14 are shown in Fig. \ref{VGG nets}.
For CIFAR-10, the propose method is scceeded to prune 85\% of parameters with only 0.53\% drop of the test accuracy. 
It is impressive that the network in which 41\% parameters are pruned achieves better test accuracy than the baseline network.
Similarly, we can prune 41\% of parameters with only 0.85\% drop of the test accuracy for CIFAR-100. For TinyImageNet-200, 37\% of parameters can be pruned with only 0.70\% drop of the test accuracy. 

For all datasets, we observe similar tendency of pruning such that the parameters in the deep layer are pruned much more than the shallow layer.

\subsection{Pruning of ResNet}

We also have performed experiments with ResNet for CIFAR-10, CIFAR-100, and TinyImageNet-200.
%
%
For CIFAR-10/100, we trained ResNet20 (19 convolution(conv) + one fully connected(fc) layers) and ResNet32 (31-conv + 1-fc layers) with batch normalization layers.
For TinyImageNet-200, ResNet18 (17 convolution(conv) + one fully connected(fc) layers) and ResNet34 (33-conv + 1-fc layers) were trained with batch normalization layers.
Parameter settings are the same with the experiments for VGG net.


\begin{figure}[tbp]
\centering
\subfloat[ResNet20 on CIFAR-10]{
\includegraphics[scale=0.28]{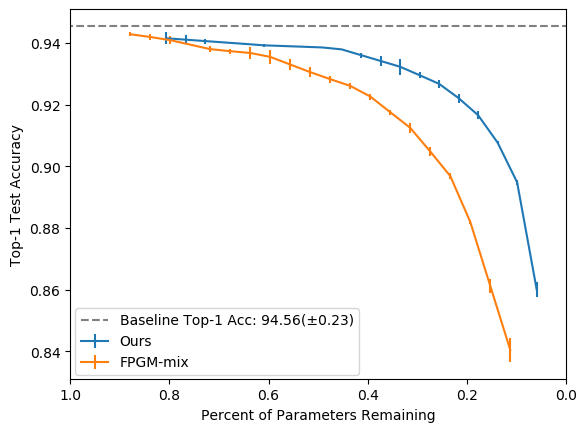}
\includegraphics[scale=0.28]{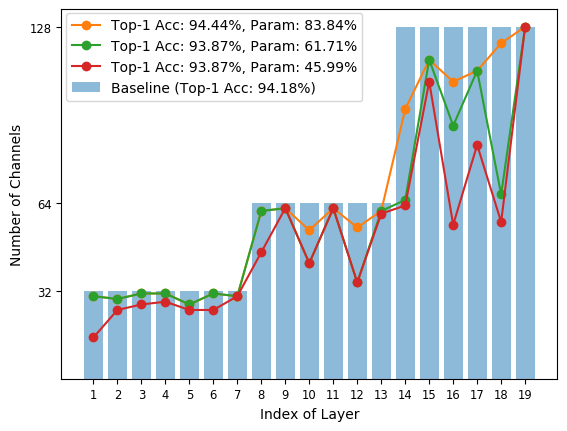}
\label{ResNet20 on CIFAR-10}}
\hfil
\subfloat[ResNet20 on CIFAR-100]{
\includegraphics[scale=0.28]{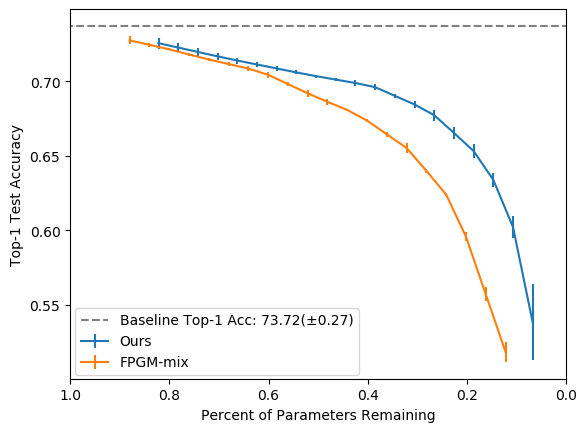}
\includegraphics[scale=0.28]{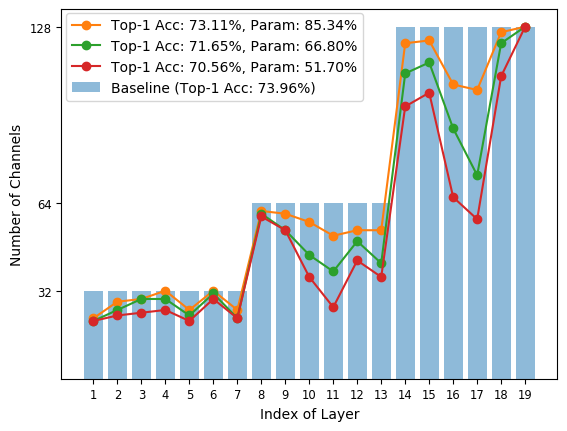}
\label{ResNet20 on CIFAR-100}}
\hfil
\subfloat[ResNet18 on TinyImageNet-200]{
\includegraphics[scale=0.28]{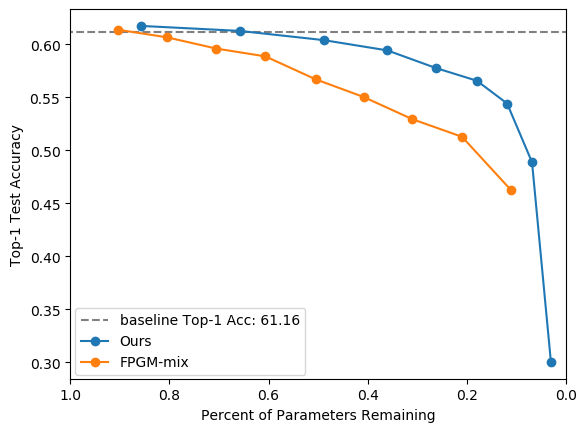}
\includegraphics[scale=0.28]{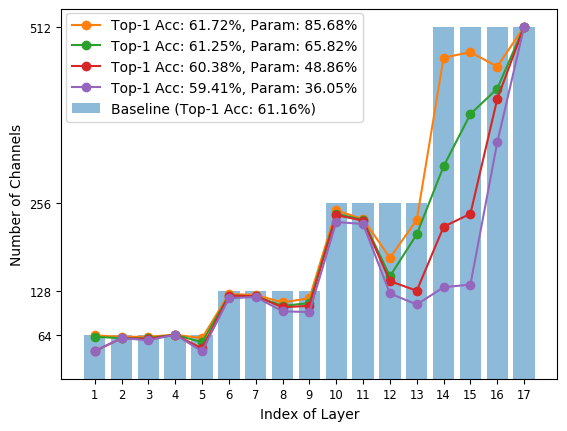}
\label{ResNet18 on TinyImageNet-200}}
\caption[]{The results with ResNet20 and ResNet18. Comparison of test accuracy of the pruned networks (left). The average of three trials is shown. The numbers of the pruned channels in each layer of the networks are shown in the right figure.
}
\label{ResNet20 and ResNet18}
\end{figure}

\begin{figure}[htbp]
\centering
\subfloat[ResNet32 on CIFAR-10]{
\includegraphics[scale=0.28]{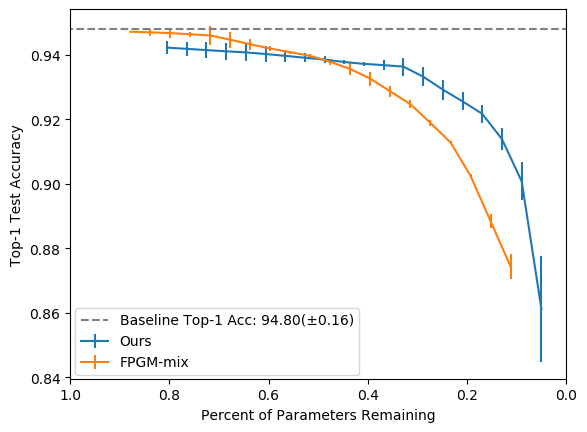}
\includegraphics[scale=0.28]{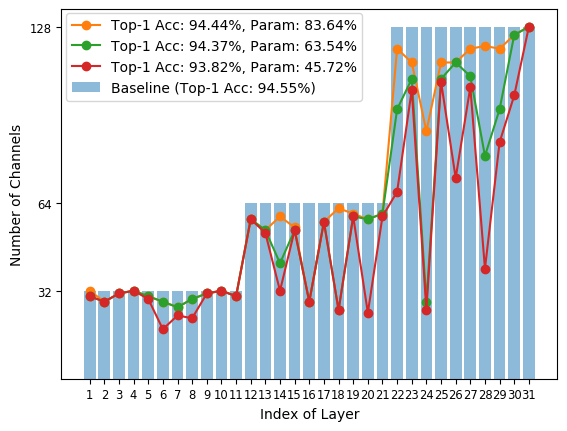}
\label{ResNet32 on CIFAR-10}}
\hfil
\subfloat[ResNet32 on CIFAR-100]{
\includegraphics[scale=0.28]{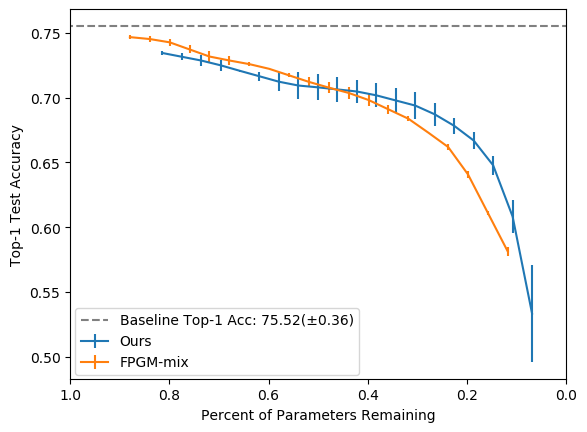}
\includegraphics[scale=0.28]{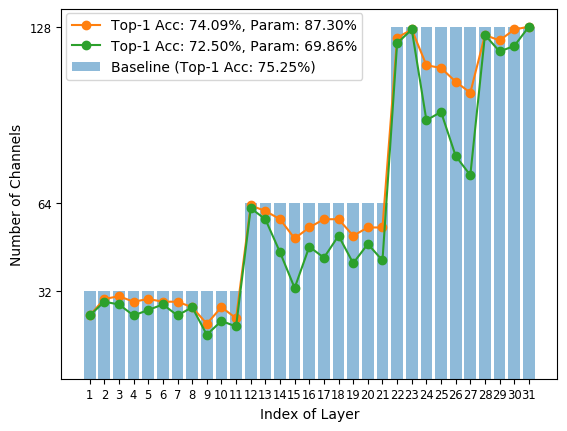}
\label{ResNet32 on CIFAR-100}}
\hfil
\subfloat[ResNet34 on TinyImageNet-200]{
\includegraphics[scale=0.28]{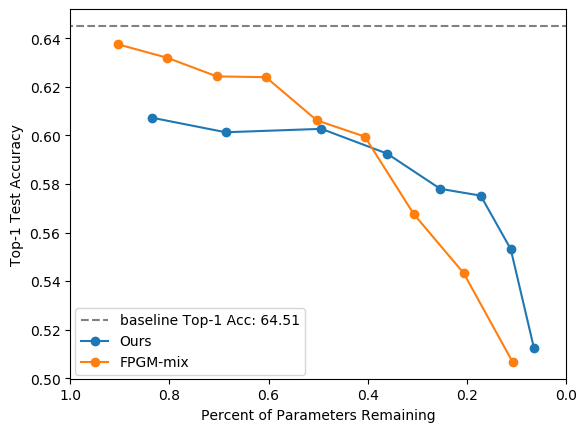}
\includegraphics[scale=0.28]{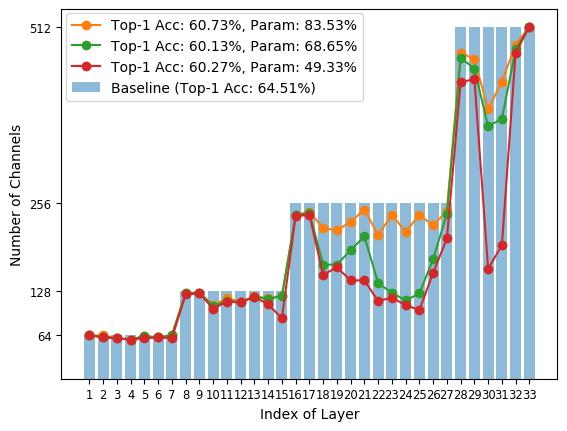}
\label{ResNet34 on TinyImageNet-200}}
\caption[]{The results with ResNet32 and ResNet34. Comparison of test accuracy of pruned networks (left). For CIFAR-10/100, the average of three trials is shown. Right figure shows the numbers of pruned channels in each layer of the networks.
}
\label{ResNet32 and ResNet34}
\end{figure}

The results with ResNet are shown in Fig. \ref{ResNet20 and ResNet18} and Fig. \ref{ResNet32 and ResNet34}.
By using the proposed pruning method, we can obtain the compact network which has only 54\% of the parameters with the baseline network but can achieve almost same test accuracy with the baseline network (only 0.31\% drop) for ResNet20 on CIFAR-10. 
The pruned network with 16\% parameters gives better test accuracy than the baseline network.
Similaly, for ResNet20 on CIFAR-100, the network with 15\% parameters gives almost same test accuracy with the baseline network (only 0.85\% drop). 
The network pruned 34\% of the parameters for ResNet18 on TinyImageNet-200 achieved higher test accuracy than the baseline network.
The network pruned 54\% of the parameters for ResNet32 on CIFAR-10 and the network pruned 13\% of the parameters for ResNet32 on CIFAR-100 are also gives the almost same test accuracy (only 0.73\% drop and only 1.2\% drop).
For ResNet32 on TinyImageNet-200, we can prune the network for 16\% of the parameters with only 3.8\% drop of the test accuracy.
From the right figures, it is noticed that the parameters in the deep layer of each residual block are pruned more.

\subsection{Comparison with The State-of-the-art Method}


The proposed method was compared with one of the state-of-the-art filter pruning methods FPGM-mix.
The filter pruning via geometric median (FPGM) \cite{he2019filter} is one of the state-of-the-art method and FPGM-mix is a mixture of FPGM and their previous norm-based method \cite{he2018soft}. 
We have performed experiments with the same pruning rate of $P$. 
The ratio of FPGM and the norm-based method is determined according to the paper \cite{he2018soft}.
Namely, $3/4$ of the filters are selected with FPGM and the remaining $1/4$ filters are selected with the norm-based criterion. 


The results of the comparisons are shown in Fig.\ref{VGG nets}, Fig.\ref{ResNet20 and ResNet18} and Fig.\ref{ResNet32 and ResNet34}.
For VGG nets, ResNet20 and ResNet18, the performance of the pruned network with the proposed method is better than the FPGM-mix for all pruning ratios.
For ResNet32 and ResNet34, the test accuracy of the pruned network with the proposed method gives better than the network obtained by FPGM-mix when the pruning rate is larger than 60\%.
Our proposed method consider the structure of the networks to prune unnnecessary parameters. So that, the proposed method gives better performance than the  FPGM-mix  when the pruning rate is high.

These results show the effectiveness of the proposed method compares to the state-of-the-art method, especially when the network is pruned more than 50\% of the parameters.


\section{Conclusion}

We propose a filter pruning method with the hierarchical group sparse regularization based on the feature-wise grouping for pruning filters, which connected to unnecessary input channels, using the influence of the classification loss.
At first the network was trained with the hierarchical sparse regularization.
We take the strategy of the step-wise pruning of the filters by searching the filter with the minimum loss increase. 
Then the obtained compact network was retrained from scratch.
Experiments using CIFAR-10/100 and TinyImageNet-200 datasets show the outstanding performance than the state-of-the-art pruning method. 
Especially the performance of the pruned network is better than the state-of-the-art pruning method when more than 50\% of the parameters are pruned.

\bibliographystyle{unsrt}
\bibliography{root.bib}

\end{document}